# An Early Prediction of COVID-19 Associated Hospitalization Surge Using Deep Learning Approach

**Yuqi Meng**
Benxi Senior High School
Benxi, Liaoning, China
1439859601@qq.com

**Qiancheng Sun**
Northeast Yucai Foreign Language School
Shenyang, Liaoning, China
someone0408@hotmail.com

**Suning Hong**
Northeast Yucai Foreign Language School
Shenyang, Liaoning, China
3343108177@qq.com

**Zhixiang Li**
Department of Biomedical Engineering
Shenyang Pharmaceutical University
Shenyang, Liaoning, China
106040205@syphu.edu.cn

## Abstract

The global pandemic caused by COVID-19 affects our lives in all aspects. As of September 11, more than 28 million people have tested positive for COVID-19 infection, and more than 911,000 people have lost their lives in this virus battle. Some patients can not receive appropriate medical treatment due the limits of hospitalization volume and shortage of ICU beds. An estimate future hospitalization is critical so that medical resources can be allocated as needed. In this study, we propose to use 4 recurrent neural networks to infer hospitalization change for the following week compared with the current week. Results show that sequence to sequence model with attention achieves a high accuracy of 0.938 and AUC of 0.850 in the hospitalization prediction. Our work has the potential to predict the hospitalization need and send warning to medical providers and other stakeholders when a re-surge initializes.

## 1 Introduction

COVID-19, the disease caused by SARS-CoV-2 virus, is affecting our daily life from financing, health, employment, travel restriction, family apart to every details [1]. According to the study of John Hopkins University [2], by September 11th, more than 28 million people have been tested positive of COVID-19 infection. We lost more than 911,000 lives in this battle of the virus. United States of America has the most serious community transmission and fatal cases. Solely, in U.S.A, more than 6 million people are infected with COVID-19, which accounts for more than 196,000 death. Until a efficient, risk-free and large-scale available vaccine is developed, we have to keep fighting with the disease and can't resume to our daily life before the global pandemic.

Most patients infected with COVID-19 only have mild symptoms such as cough, fever, muscle pain and so on [3]. These patients could recover with self-quarantine and medical treatment by their patients. However, people aged over 65 or having underlying medical condition such as lung disease, chronic disease, diabetes, are more at risk for complications from the virus [4]. Those patients with serious symptoms should receive hospitalization under the guidance of medical provider. According to the data of US Centers for Disease Control and Prevention, the overall cumulative hospitalization rate associated with COVID-19 is 166.9 per 100,000 with the highest rates in people aged 65 years and older (451.2 per 100,000) followed by the age group of 60-64 years (249.8 per 100,000). A high hospitalization rate may cause the shortage of health resources, higher risk of community transmission and extra burden to health provider. Therefore, a precise estimate of surge of hospitalization is essential.

Numerous models are being used to predict the hospitalization rates for the following four weeks [5]. These models can be formed to 2 groups. In one group, machine learning and deep learning are widely used to infer the trend of hospitalization change. In the other group, traditional statistics, such as naive estimator and auto regression model are being applied to make predictions. In this work, we will predict the hospitalization change of the next one week from multi-modal data using different deep learning configurations.

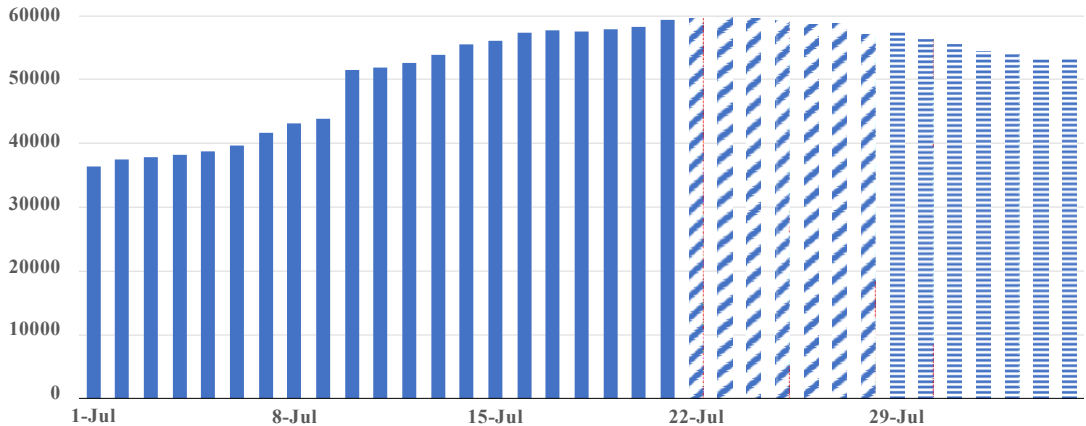

Figure 1: 35 days running hospitalization of U.S.A starting from July 1st, 2020

## 2 Related Work

Thanks to the fast development of computational technology, deep learning and machine learning has been widely used in the different aspects. Healthcare also benefits a lot from many deep learning and machine learning models built by researchers including performance improvement of disease early prediction [6, 7], automatic interpretation of medial images [8, 9], RNA sequence pattern identification [10] and so on. In terms of epidemic prediction, time series deep learning models are also widely involved. Ning et al [11]. apply a 2-step augmented regression to estimate regional influenza epidemics. Kondo et al. [12] use sequence-to-sequence model to achieve state-of-the-art Pearson score in the influenza prevalence prediction. Particularly to COVID-19, the Karlen Working Group uses the rate of reported infections to estimate the number of new hospitalizations in a given jurisdiction. The Institute of Health Metrics and Evaluation infers numbers of new hospitalizations based on numbers of predicted deaths. Unlike these work, we will use statistical figure of the disease of the past week to infer whether the hospitalization of the following week will surge compared with current week. This work will contribute to the medical resource allocation and disease prevention.

## 3 Methods

### 3.1 Dataset

The dataset we use is named COVIDTracking and is published at https://github.com/google-research/open-covid-19-data/. 20 epidemiological statistics of U.S.A including positive cases, death, test number, cumulative hospitalization and other are collected from January 1st, 2020. The dataset is updated everyday. An example of 35 days running hospitalization is shown in Figure 1. Our research problem is defined as follows. We are trying to infer the average hospitalization change from the epidemiological statistics of past 4 weeks. The average hospitalization change is a binary label and defined as the comparison between averaged next 7 days running hospitalization and averaged past week (including today) running hospitalization. We used Figure 1 to illustrate our task. Suppose the reference date is July 28th, 2020. The input features are 20 epidemiological statistics from July 1st to July 28th, 2020. The predicted outcome is 0 in that the running averaged hospitalization from July 29th to August 4th (bars with horizontal strips) is smaller than that of July 22th to July 28th (bars with diagonal strips). As the cases number and other factors before March 1st are highly limited in U.S. Therefore, we will only use the data after March 1st. Given that our time lag is 4 weeks in the prediction, our reference date is from April 1st, 2020 to August 30th, 2020. There are 152 days in total.

Table 1: Performance of 4 recurrent neural networks on the prediction of hospitalization change associated with COVID-19

| Model | Accuray | AUC |
|---|---|---|
| Long Short-Term Memory | 0.639 | 0.539 |
| Stacked Long Short-Term Memory | 0.738 | 0.773 |
| Bi-directional Long Short-Term Memory | 0.885 | 0.767 |
| Sequence-to-sequence Model with Attention | 0.934 | 0.850 |

### 3.2 Deep Learning Models

Recurrent neural networks (RNN) are widely used in the time series prediction. In our study, we will use a single layer of Long short-term memory (LSTM) [13] with 64 neurons as the baseline. On top of this we will try some other configurations with augmentation and attention. First, we will try a 3-layer stacked LSTM model [14] with 128, 64 and 32 neuron at each LSTM layer. Next, we adopt Bidirectional LSTM (Bi-LSTM) [15]. The bidirectional structure allows the network to transmit both forward and backward at every time step. We also try sequence-to-sequence model with attention. In this model, LSTM layers are used as both encoder and decoder. This model yields state-of-the-arts model in influenza prediction. We expect it can also boost the performance of naive LSTM models. For the technical details we will use stochastic gradient descent optmizier with learning rate 0.01 and a momentum of 0.9. As our problem is a binary classification, we employ binary cross-entropy as our loss function. For the stacked LSTM and the LSTM components of seq-to-seq model, the LSTM layer is followed by a rectifier activation and dropout layer with dropout rate 0.2. We will train each classifier for 200 epochs with early termination conditioned on the loss of developing set.

### 3.3 Evaluation

We first split all data into training set and test set. In the training set, the reference date is from April 1st to Jun 30th, 2020 (91 days in total). In the test set, the reference date is from July 1st to August 30th, 2020 (61 days in total). We randomly select 10 days from training set as a development set for hyper-parameters tuning purposes. We evaluate the held-out test set using accuracy and area under the receiver operating characteristic (AUC).

### 3.4 Implentation

We build the entire pipeline using Python 3.6. Keras is used to design deep neural networks. Pandas and Numpy is applied to pre-process data. Google Colab with GPU runtime is used to reduce training time. Scikit-learn package is used to evaluate the model performance.

## 4 Results

The results are shown in Table 1. Sequence to sequence model with attention yields the best accuracy of 0.934 and AUC of 0.850, followed by bidirectional LSTM, which achieves an accuracy of 0.885 and AUC of 0.767. We would like to note that the test sample is only 61. Bi-LSTM correctly predict 56 out of 61 hospitalization change while Seq2seq outperforms a bit that correctly infer one more hospitalization change. LSTM model and stacked LSTM, due to its simple architecture, don't achieve optimal results.

## 5 Discussion and conclusion

We visualized the predicted trend compared with ground truth. We can see that LSTM model simply predicts all the time series to negative. This is as expectation as the simple LSTM network will cause over-fitting on the training set. In the real case, running hospitalization change flip over from increaing to decreasing on July 22nd, 2020. Both Seq2seq model and Bi-LSTM could give an estimate of this flipping over trending. When talking about the practical use, our time series classifier could give an estimate of the hospitalization change. This may help the health provider and stakeholder to allocate the limited medical source. Moreover, it can also send warning to people when there is a COVID-19 re-surge.

Our work also have following limitations. First, as COVID-19 is caused by a novel virus and only traced recently, we only have a sample size of 152. A larger dataset could potential improve our model performance. Secondly, we don't apply some other trending time series prediction model such as deep transformers [16], which has shown great success in influenza prediction [17]. Finally, we don't try other configurations such as 2 week time lag, the monthly hospitalization change and so on. In the future study, we will keep monitoring this global pandemic and update as well as upgrade our model to achieve better hospitalization change prediction.

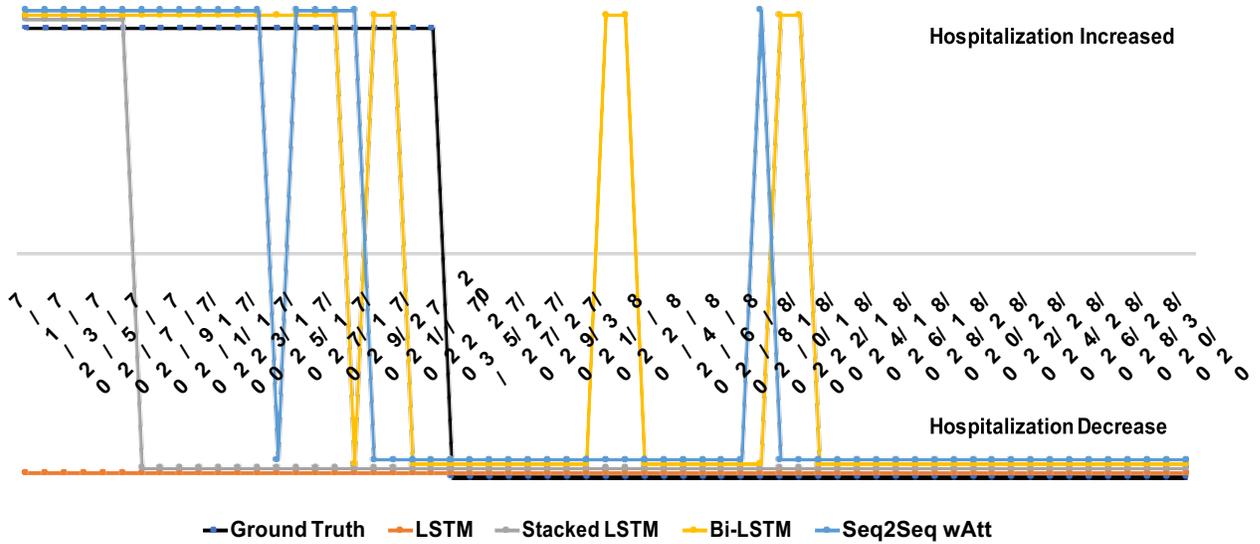

Figure 2: Predicted trends from 4 recurrent neural networks compared with the ground truth trend

## 6 Author Contribution Statement

All authors discuss and define the research problem. Y.M. extracts and preprocess all the research data. Y.M. and Y.L. implement all deep neural networks, analyze and visualize the results. Y.M. and Y.L. write the manuscript advised by Z.L., who also organizes the manuscript to a better academic standard.